\documentclass[preprint,12pt]{elsarticle}

\usepackage{graphicx}

\usepackage{amssymb}
\usepackage{acronym}
\usepackage{ctable}
\usepackage{multirow}
\usepackage{lscape}
\usepackage{array}

\DeclareGraphicsExtensions{.eps}

\journal{Computer Vision and Image Understanding}

\begin{document}

\begin{frontmatter}


\author{Ana Paula Brand\~ao Lopes$^{1,2}$}
\ead{paula@dcc.ufmg.br}
\author{Eduardo Alves do Valle Jr.$^{3}$}
\ead{mail@eduardovalle.com}
\author{Jussara Marques de Almeida$^{1}$}
\ead{jussara@dcc.ufmg.br}
\author{Arnaldo Albuquerque de Ara\'ujo$^{1}$}
\ead{arnaldo@dcc.ufmg.br}
\address{$^{1}$Depart. of Computer Science -- Universidade Federal de Minas Gerais (UFMG)\\ Belo Horizonte (MG), Brazil\\
$^{2}$Depart. of Exact and Tech. Sciences -- Universidade Estadual de Santa Cruz (UESC)\\ Ilh\'eus (BA), Brazil\\
$^{3}$Institute of Computation -- Universidade Estadual de Campinas (UNICAMP)\\ Campinas (SP), Brazil}

\title{Action Recognition in Videos:\\ from Motion Capture Labs to the Web}

%
%

\begin{abstract}
This paper presents a survey of human action recognition approaches based on visual data recorded from a single video camera. We propose an organizing framework which puts in evidence the evolution of the area, with techniques moving from heavily constrained motion capture scenarios towards more challenging, realistic, ``\emph{in the wild}'' videos. The proposed organization is based on the representation used as input for the recognition task, emphasizing the hypothesis assumed and thus, the constraints imposed on the type of video that each technique is able to address.  Expliciting the hypothesis and constraints makes the framework particularly useful to select a method, given an application. Another advantage of the proposed organization is that it allows categorizing newest approaches seamlessly with traditional ones, while providing an insightful perspective of the evolution of the action recognition task up to now. That perspective is the basis for the discussion in the end of the paper, where we also present the main open issues in the area.
\end{abstract}

\end{frontmatter}
\newacro{APM}[]{Public Archives of Minas Gerais -- \textit{Arquivo P\'ublico Mineiro}}
\newacro{BoVF}[BoVF]{Bag of Visual Features}
\newacro{BoW}[BoW]{Bag-of-Words}
\newacro{CFV}[CFV]{Contextual Feature Vector}
\newacro{CBIR}[CBIR]{Content-Based Image Retrieval}
\newacro{CBVR}[CBVR]{Content-Based Video Retrieval}
\newacro{DoG}[DoG]{Difference of Gaussians}
\newacro{DTW}[DTW]{Dynamic Time Warping}
\newacro{EMD}[EMD]{Earth Mover's Distance}
\newacro{GMM}[GMM]{Gaussian Mixture Models}
\newacro{HCI}[HCI]{Human-Computer Interaction}
\newacro{HMM}[HMM]{Hidden-Markov Models}
\newacro{HoF}[HoF]{Histograms of optical Flow}
\newacro{HoG}[HoG]{Histograms of Gradients}
\newacro{IR}[IR]{Information Retrieval}
\newacro{KTH}[KTH]{Royal Institute of Technology}
\newacro{LDA}[LDA]{Latent Dirichlet Allocation}
\newacro{LIBSVM}[LIBSVM]{LIBrary for Support Vector Machines}
\newacro{LSCOM}[LSCOM]{Large-Scale Concept Ontology for Multimedia}
\newacro{MEI}[MEI]{Motion Energy Image}
\newacro{ML}[ML]{Machine Learning}
\newacro{MHI}[MHI]{Motion History Image}
\newacro{MMI}[MMI]{Maximization of Mutual Information}
\newacro{MPEG}[MPEG]{Moving Picture Experts Group}
\newacro{NMF}[NMF]{Non-negative Matrix Factorization}
\newacro{NN}[NN]{Neural Networks}
\newacro{PCA}[PCA]{Principal Component Analysis}
\newacro{pLSA}[pLSA]{probabilistic Latent Semantic Analysis}
\newacro{RBF}[RBF]{Radial-Basis Function}
\newacro{ROI}[ROI]{Region of Interest}
\newacro{RVM}[RVM]{Relevance Vector Machines}
\newacro{SFF}[SFF]{Smart Fast-Forward}
\newacro{SIFT}[SIFT]{Scale-Invariant Feature Transform}
\newacro{SSM}[SSM]{Self-Similarity Matrix}
\newacro{ST-SIFT}[ST-SIFT]{Spatio-Temporal Scale-Invariant Analysis}
\newacro{STIP}[STIP]{Space-Time Interest Points}
\newacro{STPM}[STPM]{Spatio-Temporal Pyramid Matching}
\newacro{SURF}[SURF]{Speed-Up Robust Features}
\newacro{SVM}[SVM]{Support Vector Machines}
\newacro{tf-idf}[tf-idf]{term-frequency inverse-document-frequency}
\newacro{TREC}[TREC]{Text REtrieval Conference}
\newacro{TRECVID}[TRECVID]{\acs{TREC} Video Retrieval Evaluation}
\newacro{VWC}[VWC]{Video Words Clusters}
\acresetall
\section{Introduction}
\label{sec:intro}
This paper presents the state-of-the-art in action recognition for videos based on visual data recorded from a single camera. It shows how the approaches have evolved from the analysis of videos produced in heavily constrained motion capture environments towards recent attempts of automatically understanding realistic or ``\emph{in the wild}'' videos. Our goal is not to provide an exhaustive survey, but rather an elucidative overview of the main ideas in the area, its historical evolution and its current trends.

A number of existing review papers more or less related to the task of action recognition in videos provide different perspectives on the field. Put together, they also provide the perspective of its evolution (Section~\ref{sec:previous-surveys}). However, most of them fail to cover the most recent achievements of the area. The few exceptions inherit the traditional organization and taxonomy of older surveys, which are unable to characterize the current corpus of methods adequately.

In this survey, a new organizing scheme including both old and recent approaches is proposed in Section~\ref{sec:approaches-categorization}. For the sake of completeness, an overview of older approaches is presented in Section~\ref{sec:object-based}, while recent approaches are covered in deeper detail in Section~\ref{sec:statistical-approaches}. A summary followed by a discussion on current research trends is provided in Section~\ref{sec:discussion}. Finally, concluding remarks are presented in Section~\ref{sec:conclu-approaches}.

\subsection{Why Should We Recognize Actions in Videos?}
In recent years, Internet users witnessed the emergence of a great amount of multimedia content in the Web. Initially, such kind of content was generated by professional or semi-professional individuals or enterprises, in a typical broadcasting scheme. However, in a second wave, the users themselves started creating and publishing their own multimedia materials. This motion towards user-generated content was motivated by a number of factors, mainly the drop in the cost of devices such as cameras and microphones and the spread of high-bandwidth connections, as well as the emergence of Web 2.0 applications, including online social networks. This new scenario led to an overwhelming increase in the amount of multimedia content available, which by its turn brought up the limitations of traditional Web tools in dealing with non-textual data.

To make multimedia data effectively available, the high-level indexing aimed at meaningful, semantically-oriented retrieval is a critical goal. While for text, the words themselves convey quite directly its semantics, in the case of visual information, the connection between low-level encoding (i.e., pixels) and semantic meaning is far from immediate. Indeed, it is an open research issue, commonly referred to as the \emph{semantic gap} \cite{smeulders2000}.

The current state-of-the-art in terms of systems for image and video retrieval is described in \cite{snoek2008}. Such systems are composed by several single individual concept detectors which are applied independently to every item in the database. The estimated probabilities of occurrence then compose the feature vectors for the search engine. Systems like the one just described have the advantage of enabling textual search, as opposed to query-by-example\footnote{In query-by-example systems, the users choose an image as a query and the system returns those ones considered most similar to it.} or query-by-sketch\footnote{In query-by-sketch systems, the users need to draw a rough sketch of the image they want to find.} approaches, which are not always intuitive for users accustomed to commercial search engines.

One key issue in such approach is which concepts should be considered. This issue was addressed in the \ac{LSCOM} workshop \cite{kennedy2006a}, which defined a lexicon containing around 1000 concepts, from which 449 have been annotated in 80 hours of video coming from the \ac{TRECVID} 2005 database \cite{over2005}.

Later on, experiments performed by \cite{kennedy2006} showed that annotations for some concepts defined in \ac{LSCOM} varied significantly whether the annotators watched video sequences or looked at keyframes. Those results indicate that the dynamic nature of video information plays an important role in the recognition of some concepts. Also, they suggested that such dynamic semantics cannot adequately be captured by the direct application of techniques aimed at still images.

The 24 activity/event \ac{LSCOM} concepts which had their annotations changed after the experiments described in \cite{kennedy2006} are listed in Table~\ref{tab:lscom-dynamic}. From that table, it is possible to see that all those concepts, either directly or indirectly, are related to actions performed by human beings.

\begin{table}[ht]
\caption{\acs{LSCOM} concepts that were found highly dependant on motion. Users frequently
re-annotated them when switching to viewing video segments instead of keyframes \cite{kennedy2006}.}
\begin{center}
\begin{tabular}{c c}
\toprule
Airplane Crash &  Greeting  \\
\midrule
Airplane Flying & Handshaking \\
\midrule
Airplane Landing & Helicopter Hovering \\
\midrule
Airplane Takeoff  &  People Crying \\
\midrule
Car Crash & People Marching  \\
\midrule
Cheering  & Riot \\
\midrule
Dancing & Running  \\
\midrule
Demonstration Or Protest  & Shooting   \\
\midrule
Election Campaign Debate &  Singing \\
\midrule
Election Campaign Greeting & Street Battle \\
\midrule
Exiting Car &  Throwing \\
\midrule
Fighter Combat & Walking \\
\toprule
\end{tabular}
\label{tab:lscom-dynamic}
\end{center}
\end{table}

\subsubsection*{Additional Applications}
Although the improvement of high-level video indexing and retrieval is an important motivation for action recognition research, it is worth mentioning several additional applications.

A great amount of work has been done around the idea of building ``smart'' video surveillance systems, which would be able to detect suspicious behavior automatically. In \cite{lavee2007}, for instance, a framework to aid the search for specific events in recorded surveillance video is proposed. In addition, recognition of people by their gait has been studied as alternative biometrics \cite{kale2004}. A review focused on visual surveillance systems is presented in \cite{hu2004}.

The analysis of sport videos is another important application. In \cite{xie2004}, for example, the classification of video segments between play and break intervals is suggested to summarize the video, by taking out the breaks. Soccer games are also analyzed by \cite{zhu2007}, in which  text and the players'  trajectories are used to build a system aimed at helping coaches in tactical analysis. Six actions of a cricket umpire are analyzed in \cite{rahman2005} -- by a technique using an appearance-based method similar to eigenspaces (commonly used in face recognition) -- whereas the usage of local motion analysis to identify different swimming styles is proposed in \cite{tong2006}.

Hand gesture recognition can be useful for a number of applications. In \cite{tan2004}, it is applied to identify which segments in lecture videos are worth transmitting in less compressed formats.  The underlying assumption of that work is that specific actions can indicate the importance of each sequence, therefore guiding a semantically-oriented compression. Automatic recognition of sign language symbols is explored in \cite{cooper2007} and \cite{buehler2009}, for example. In \cite{sundaram2009}, the recognition of hand manipulations of objects recorded by a camera attached to a person body are suggested as a means of interaction with a virtual reality system.

\ac{HCI} systems can also benefit from the ability of recognizing generic actions, as it can be seen in the pioneer work of \cite{bobick1999}. They present \emph{KidsRoom}, an environment able to interpret and react to specific actions of a group of children in a closed space. In a similar application, \cite{ren2002} proposed a system called \emph{smart classroom}, where the actions performed by a teacher are recognized to allow automatic camera motion and a virtual mouse. Facial actions have been recently explored either as a tool to enhance \ac{HCI} -- as in \cite{tsalakanidou2009} -- or to analyze the affective behavior of psychiatric patients \cite{cohn2009}.

A specific instance of the general content-based retrieval idea is what is called \ac{SFF}, as proposed by \cite{zelnik-manor2006}, in which the query video segment is compared against other segments in the same video in order to find similar actions taking place in different intervals.

Action recognition is an important issue also in robotics, in which the interpretation of human actions can be used either for reaction to the recognized action (i.e., control) or for learning and imitation \cite{kruger2007}.

Finally, in the medical area, human motion analysis can aid diagnosis of motor problems by comparing patient motion to normality patterns, as in \cite{albu2007}, for example. Another possible medical application is to provide remote assistance to elderly people, as suggested in \cite{kosta2008}. Similar medical applications also motivate the work of \cite{messing2009}. 
\section{Related Surveys}
\label{sec:previous-surveys}
An extensive survey on earlier studies about motion-based recognition was first presented in \cite{cedras1995}. For the authors of that work, the first step in motion-based recognition is the extraction of motion information from a sequence of images, which can be done by optical flow or motion correspondence. Motion correspondence is established by tracking specific points of interest through frames, and generating motion trajectories, which can be parameterized in several ways. Instead of computing motion information from the entire image or from specific points, region-based motion features can also be extracted. Explicit human body models are used to guide the tracking step.

The survey presented in \cite{aggarwal1997} is devoted to human motion analysis which, for them, comprises the following overall steps: a) segmentation; b) joint detection; and c) identification and recovery of 3D structures from 2D projections. The authors characterize body structure analysis as either model-based or non-model-based, depending on whether or not an \emph{a priori} shape model is used. The \emph{a priori} models considered can be stick figures, contours or spatio-temporal volumes. The proposals reviewed in \cite{aggarwal1997} are also split into two broad categories -- more related to action modeling and recognition steps (see Figure~\ref{fig:steps} in Section~\ref{sec:approaches-categorization}) -- which are: \emph{space-state models}, in which each static posture is considered as a state and state transitions occur with certain probabilities; and \emph{template matching models}, where a template is computed for each action, and then a nearest-neighbor classification scheme is applied to recognize similar actions.

While the authors in \cite{aggarwal1997} focus on approaches based on models of the human body, a survey specifically focused on models for recognition of hand gestures is presented in \cite{pavlovic1997}.

By their turn, \cite{gravila1999} reviews approaches modeling either the whole body or the hand. Selected papers are organized into three categories: \emph{2D approaches without explicit shape models}, \emph{2D approaches with explicit shape models} and \emph{3D approaches}. This survey also provides (in its Table~1) a comprehensive list of applications envisioned at the time, organized into five groups: virtual reality, smart surveillance systems, advanced user interfaces, motion analysis and model-based coding. It is worth noticing that most of the prospective applications suggested by \cite{gravila1999} still remain as unsolved challenges.

In \cite{shah2003}, the recognition of human action is described as comprising the following -- more general, in comparison to \cite{aggarwal1997} -- steps: a)~extraction of relevant visual information; b)~representation of that information in a suitable form; c) interpretation. The specific modeling of human body or body parts is not seen by the author as an essential step for human action recognition. In contrast, tracking and trajectory computation are considered the primary subtasks. Therefore, this survey is focused specifically on trajectory-based techniques.

For \cite{wang2003}, human motion analysis comprises the following steps: a)~motion segmentation; b)~object classification of segmented moving regions -- which can be shape or motion-based; c)~tracking of identified objects along consecutive frames; and d)~recognition of motion patterns, providing what they call \emph{behavior understanding}. As in \cite{aggarwal1997}, action modeling approaches are distinguished between \emph{template-based} and \emph{space-states based}.

The authors of \cite{buxton2003} present a survey from the perspective of the generative learning algorithms applied to any of the various processing steps of action understanding systems. In that paper, such systems are categorized generically into \emph{explicit models} and \emph{exemplar-based models}.

In \cite{wang2003b}, the focus falls again onto approaches relying on human-body or body parts. That survey is mainly motivated by biometrics applications, and the paper is composed of two main parts: in the first, the author provides a detailed survey on tracking techniques applied to heads, hands or the whole body. In the second, techniques for analyzing different models for those tracked elements are reviewed.

The work of \cite{aggarwal2004} expands and updates the earlier review presented in \cite{aggarwal1997}, by including not only actions, but also interactions. The surveyed approaches are distinguished by the level of detail in which the moving objects are described. \emph{Coarse level approaches} are those in which people are considered as bounding boxes or ellipses. Then, motion patterns are used to model the actions. In approaches lying in the \emph{intermediate level} of detail, people are represented by large body parts or silhouettes. Finally, \emph{detailed models} can be built on the entire body or on specific parts, such as hands in the case of gesture recognition tasks.

Action modeling approaches are also distinguished based on two different aspects: the first differentiates \emph{direct recognition} from reconstruction of \emph{body models} before recognition. The second aspect distinguishes approaches by their \emph{static} or \emph{dynamic nature}, if the recognition is performed on a frame-by-frame basis or taking the entire sequence as the basic unit analysis. High-level recognition schemes -- similar to those which \cite{wang2003} call behavior understanding -- are also discussed, most of them relying on manually constructed semantic models of the world.

In \cite{hu2004} an extensive review on papers related to surveillance systems is provided. The authors consider a visual surveillance system comprising of the following steps: a)~motion detection, which includes object modeling, segmentation and classification; b)~tracking of moving objects; and c)~behavior understanding. For some applications, an additional step of natural language description can be added. The ability to identify people at a distance (gait-based recognition) can also be introduced.

In another review focused on trajectory-based approaches, \cite{chellappa2005} define what it is called \emph{activity inference}, comprised, in their view, of three steps: a) low-level video processing; b) trajectory modeling; c) similarity computation. In this scheme, low-level processing is aimed at computing trajectories for selected objects. The trajectories for each action can be modeled by varied techniques and for each model a similarity measure needs to be established.

The survey of \cite{moeslund2006} offers an overview of human motion analysis in general, with a section devoted to action recognition. They suggest that action recognition approaches can be broadly separated between the ones that explicitly consider human presence in the scene and the ones that do not. The recognition section of that paper is structured around three different kinds of tasks: \emph{scene interpretation}, without identifying particular objects; \emph{holistic recognition}, using the human body or body parts, to recognize both the subjects and the actions performed by them; \emph{action primitives and ``grammers"}, in which motor primitives are used for representation or control. The primitives in the latter task can be used to create an action hierarchy that gives a semantic description of the scene. However, in most of such approaches motion primitives are usually taken as already available.

The review of \cite{kruger2007} is focused on robotics applications, more specifically for learning and imitation. They distinguish approaches based on: \emph{scene interpretation}, in which the moving objects are not ``identified", but have only their overall motions analyzed; the \emph{body as a whole}; \emph{body parts} and \emph{grammars}.

The review presented in \cite{poppe2007} deals specifically with pose estimation, assumed as a needed step for action recognition.

In the recent short review presented in \cite{ahad2008}, which is explicitly focused on papers from 2001 to 2008, a hierarchical terminology composed of \emph{action primitives}, \emph{actions} and \emph{activities} is adopted. This survey categorizes different proposals according to the \ac{ML} techniques applied, regardless of the underlying representation.  In other words, it is focused on the modeling step (Figure~\ref{fig:steps}).

The major branches presented in \cite{turaga2008} differentiate between approaches aimed at \textit{actions} and those aimed at \textit{activities} recognition. In their case, similarly to \cite{moeslund2006}, actions are defined as simple motion patterns executed by a unique human, while activities are more complex patterns, normally involving more than one person. The following four major steps for action recognition are identified: a)~collecting input video; b)~extracting low-level features; c)~extracting mid-level action descriptions; and d)~high-level semantic interpretations.

The low-level features considered in that survey are \textit{optical flow}, \textit{point trajectories}, \textit{blobs and shapes separated from the background} and \textit{filter responses}. According to them, actions can be described at mid-level by \textit{non-parametric}, \textit{volumetric} and \textit{parametric} models. Actions and activities can be modeled either by \textit{graphical models}, \textit{syntactic grammars-like approaches} or \textit{knowledge/logic based approaches}.

The work of \cite{ren2009} reviews motion recognition approaches in the context of \ac{CBVR}. Two major approaches are identified. In \emph{trajectory-to-trajectory approaches}, motion trajectories are extracted and compared for recognition; the category of approaches that take into account the internal structure of the object over time are denominated \emph{sequence-to-sequence approaches}.

In \cite{poppe2010}, only papers aimed at recognizing full body actions are taken into account. Image representations are separated into three large groups:  \textit{global}, when a specific \ac{ROI} is described globally, \textit{local}, based either on interest points and densely sampled ones, and \textit{application specific}. In his view, action classification can be performed either by \textit{direct classification}, using the information coming from all the frames in the sequence together and \textit{temporal state-space models}, in which action sequences are broken in smaller steps.

\section{Categorizing Different Approaches for Action Recognition}
\label{sec:approaches-categorization}
\begin{figure*}[h]
\centering
\includegraphics[width=0.9\textwidth]{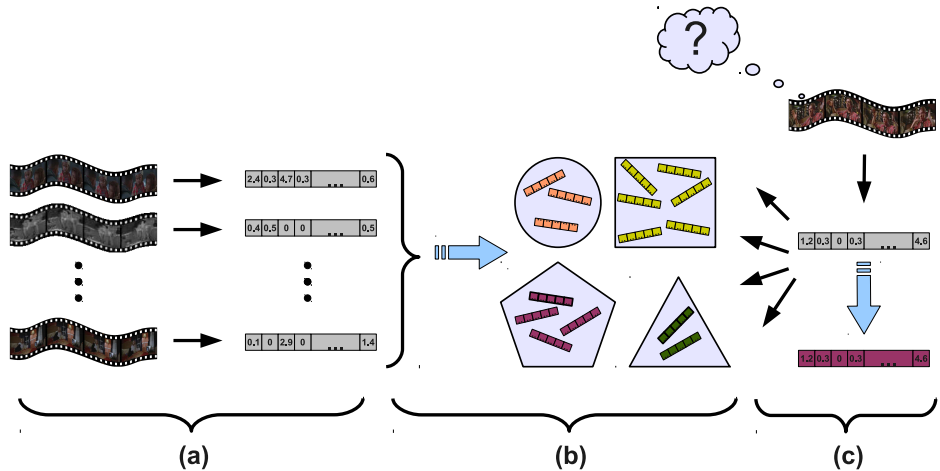}
\caption{Overview of the processing steps needed for action recognition in videos: (a) \textit{representation extraction} step, (b) \textit{action modeling} step, and (c) \textit{action recognition} step (picture best viewed in color).}
\label{fig:steps}
\end{figure*}

Regardless of the application envisioned, the process of recognizing human actions from videos can be seen as comprising the three major steps, as depicted in Figure~\ref{fig:steps}.

\begin{enumerate}[(a)]
\item  \textbf{\emph{Representation Extraction}}:  this step starts with the extraction of low-level features from the videos, like color, texture and optical flow, for example. Those features are usually fed to a somewhat complex processing chain until a suitable (i.e., compact and descriptive) representation is achieved. It is worth noticing that, unlike \cite{turaga2008}, for instance, the output of this step is the final video representation which is used as the input to the action modeling step (below), regardless of the abstraction level. This generic definition is then applied to the finer-grained hierarchical structure proposed later in this section.

\item \textbf{\emph{Action Modeling}}: in this step, the representations built in the previous step are mapped into different action categories. The \emph{spectrum} of modeling alternatives goes from the selection of a small number of action templates aimed at direct comparison to sophisticated modeling schemes involving \ac{ML} techniques.

\item \textbf{\emph{Action Recognition}}: this last step takes place when unlabeled (query) videos are analyzed against the previously built action models, so that those videos can be associated with one of the possible action categories (i.e., classified).
\end{enumerate}

As expected, those three steps are tightly interconnected. Some representation choices are more suitable for -- or are even designed specifically to go with -- certain kinds of techniques for action modeling. In the same way, the selected action modeling technique will determine -- in some cases, in a unique way -- how the classification is going to be performed.

The structure proposed in this section for organizing different approaches for action recognition is based on the representation step depicted in Figure~\ref{fig:steps}. Such choice is justified by the fact that the process of extracting a specific representation for videos is directly related to a number of assumptions about the scene content. Such assumptions, by their turn, impose specific constraints on the types of videos that each recognition approach is able to cope with. Hence, an organization of different approaches which is built based on the selected representation provides a better ground to understand the strengths and limitations of each category of approaches, making it easier to: a) select appropriate approaches for specific applications, and b) distinguish which approaches are truly comparable among them. Finally, the selected \textit{criteria} based on the underlying representation allow for sensibly unraveling unrelated approaches that end up mixed together under other categorization schemes.

Regardless of all the existing surveys discussed in Section~\ref{sec:previous-surveys}, authors follow no standard categorization structure while referring to previous papers that are related to their proposed approaches. For instance, in \cite{ebadollahi2006}, authors propose a broad categorization between \emph{object centric} and \emph{statistical} approaches, while in \cite{laptev2007a}, authors distinguish among approaches based on \emph{3D tracking of different points} of human bodies, \emph{accurate background subtraction}, \emph{motion descriptors on regions of actions} and \emph{learning of actions models}. In \cite{ke2007b}, human action approaches are categorized into those based on \emph{tracking}, \emph{flow}, \emph{spatio-temporal shapes} and \emph{interest points}. In \cite{zhao2008}, different approaches are categorized as \emph{model-based}, \emph{spatio-temporal template-based} and \emph{bags-of-visual-features-based}. In the work of \cite{wang2008}, previous papers are coarsely classified according to their specific goals, distinguishing among approaches dealing with \emph{unusual event detection}, \emph{action classification} and \emph{event recognition}.

\begin{figure}[p]
\centering
\includegraphics[width=1\textwidth]{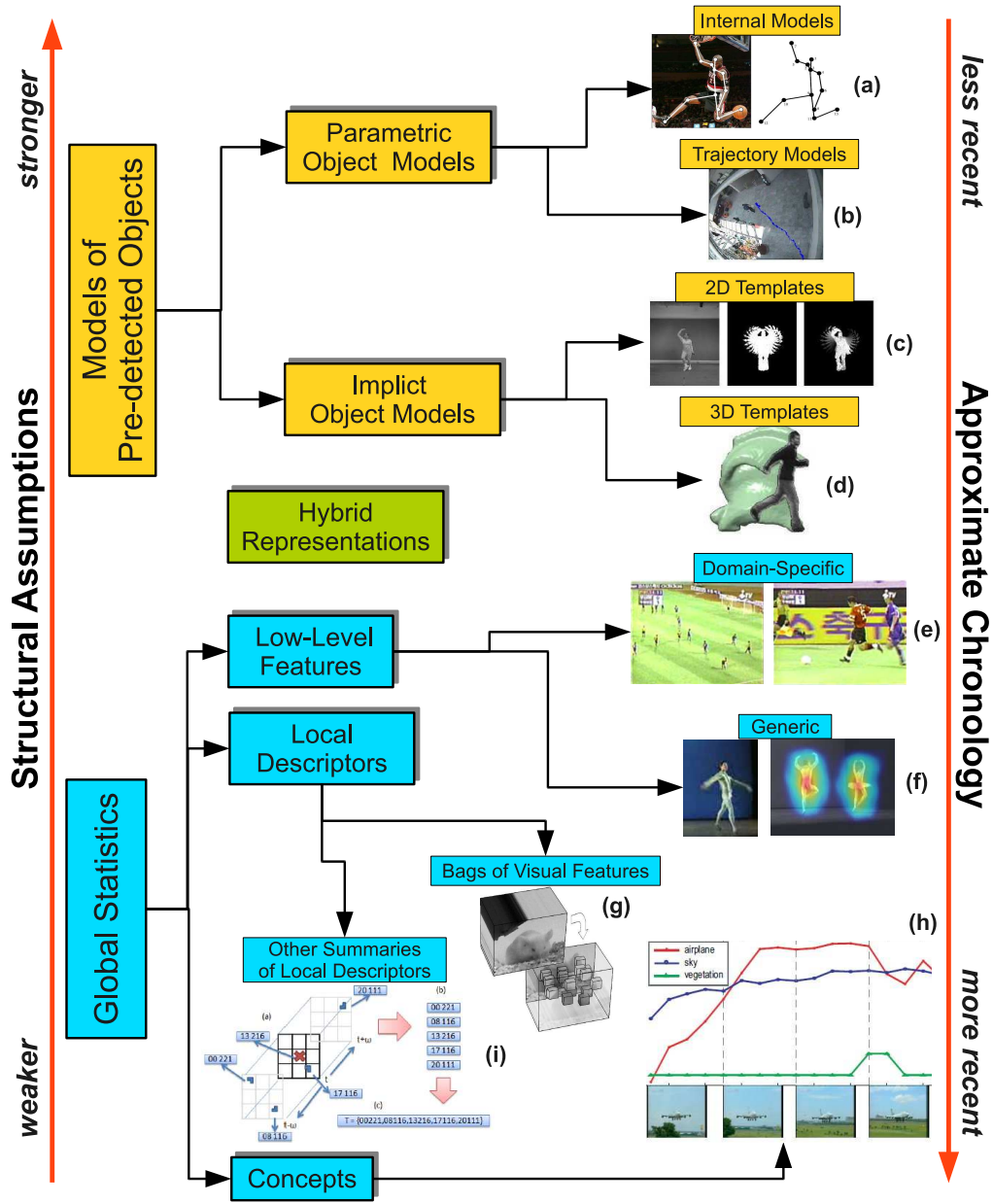}
\caption{Categorization framework used along this survey for organizing the approaches for human action recognition found in the literature. It is based on the underlying video representations. Image references: (a)~\cite{yilmaz2005}, (b)~\cite{chen2008}, (c)~\cite{bobick2001}, (d)~\cite{blank2005}, (e)~\cite{xie2004}, (f)~\cite{shechtman2005}, (g)~\cite{dollar2005}, (h)~\cite{ebadohalli2006}, (i)~\cite{gilbert2008} (picture best viewed in color).}
\label{fig:approaches}
\end{figure}

The categorization scheme we propose in this paper is depicted in Figure~\ref{fig:approaches}. In a coarse level, the different approaches are split into two large groups, which are nearly equivalent to the object centric \textit{versus} statistical categorization proposed by \cite{ebadollahi2006}. It can also be considered a generalization of the categorization of \cite{moeslund2006} into approaches that either consider the presence of humans or not. In fact, the framework proposed in this paper is a refinement of the model-based \textit{versus} model-free categorization presented in \cite{lopes2009d}, although the terms \textit{model-based} and \textit{model-free} are abandoned in order to avoid confusion between action modeling and object modeling, the former being always present (Figure~\ref{fig:steps}). The proposed scheme stresses the distinction between approaches that explicitly assume the presence of moving objects under specific conditions -- like, for example, homogeneous background -- from those in which such explicit assumption is not found.

As it will be seen, there is a non-negligible correlation between the proposed taxonomy and the temporal evolution of approaches: more recent approaches tend to rely on less constrained assumptions and therefore, more general hypothesis.

The two initial categories are further refined into subcategories organized according to the underlying representation. The vast majority of proposed solutions to human action recognition to date lies in the first broad group of approaches depicted in Figure~\ref{fig:approaches}. In other words, the video representation used by them explicitly assumes that one or more moving objects appear in the scene, typically under a number of specific conditions, like stable backgrounds and fixed scales, for example.

The basic idea behind those approaches is that it is possible to infer the actions being performed by studying the structure and/or the dynamics of the moving objects in the scene (or their parts). Moving objects of interest can be the human body, some body parts or other objects related to the application domain, like airplanes and automobiles, for example. Unlabeled moving regions can also be considered. In order to be able to analyze the moving objects, they need to be detected (and often, also tracked) before any further processing. Once the object has been detected/tracked, it can be either a)~adjusted to some pre-defined model of the object, characterized by a number of parameters (parameterized object models), or b)~characterized by global descriptors computed on their segmented area (implicit object models). Approaches relying on the presence of specific moving objects in the scene are further discussed in Section~\ref{sec:object-based}.

More recently, a number of approaches that do not explicitly rely on the presence of any specific object in the scene have been proposed. They are based on global statistical computations over different kinds of representations, within distinct abstraction levels: low-level features, mid-level interest points and high-level concepts. Approaches based on global statistics are represented by the lowest large rectangle in Figure~\ref{fig:approaches} (which starts the branch in turquoise). As such approaches have only recently been proposed, previous surveys, discussed in Section~\ref{sec:previous-surveys}, do not cover them in much detail. We provide a more comprehensive discussion of those approaches in Section~\ref{sec:statistical-approaches}.

Finally, hybrid approaches, which mix ideas from both those ones based on models of pre-detected objects and those ones based global statistics can also be found.

Tables~\ref{tab:summary-papers1},~\ref{tab:summary-papers2}~and~\ref{tab:summary-papers3} summarize the approaches for action recognition discussed throughout this paper, pointing out the main assumptions for each category, and citing related papers that are going to be discussed in the following sections.

%
%
\definecolor{repcolor}{rgb}{0.7, 0.13, 0.13}
\definecolor{assumpcolor}{rgb}{0.1, 0.1, 0.44}
\begin{table}[ht]\footnotesize
\caption{Summary of Action Recognition Approaches Based on Object Model Representations (table best viewed in color).}
\begin{center}
\begin{tabular}{m{0.001\textwidth} m{0.2\textwidth} m{0.2\textwidth} m{0.2\textwidth} m{0.2\textwidth} m{0.001\textwidth}}
\toprule
\multicolumn{6}{c}{\textbf{\textcolor{repcolor}{Basic Representation -- Object Models}}}
\\
& \multicolumn{4}{c}{\textit{\textcolor{assumpcolor}{Main Assumption: Actions can be derived from a specific model of the related objects.}}} &
\\
\multicolumn{6}{c}{[Related Papers by Subcategory (below)]}
\\
\midrule
& \multicolumn{2}{c}{\textcolor{repcolor}{Parametric Models}} & \multicolumn{2}{c}{\textcolor{repcolor}{Internal Models}} &
\\
 & \multicolumn{2}{m{0.4\textwidth}}{\centering{\textcolor{assumpcolor}{\textit{Objects related to actions obey a predefined model.}}}}
 & \multicolumn{2}{m{0.4\textwidth}}{\centering{\textcolor{assumpcolor}{\textit{Global representations of objects internal areas implicitly define the object model.}}}} &
 \\
 &
\centering{\textcolor{repcolor}{Internal Models}}
&
\centering{\textcolor{repcolor}{Trajectories}}
&
\centering{\textcolor{repcolor}{2D Descriptors}}
&
\centering{\textcolor{repcolor}{3D Descriptors}} &
\\
  &
\centering{\textcolor{assumpcolor}{\textit{A predefined object model must describe objects internal states.}}}
  &
\centering{\textcolor{assumpcolor}{\textit{The relevant information is not in the objects internal state, but in their positions over time.}}}
&
 \centering{\textcolor{assumpcolor}{\textit{Objects appearance information is enough for action recognition OR motion can be aggregated in 2D representations.}}}
  &
 \centering{\textcolor{assumpcolor}{\textit{Changes in appearance over time are also relevant for action recognition.}}} &
 \\
\midrule
& \centering{\cite{yilmaz2005} \cite{peursum2004} \cite{peursum2005} \cite{jiang2006b} \cite{kosta2006} \cite{filipovych2008} \cite{wang2009} \cite{dhillon2009}}
&
\centering{\cite{chen2008} \cite{abdelkader2008} \cite{cuntoor2008a} \cite{hu2008a} \cite{hu2008b}}
&
\centering{\cite{bobick2001} \cite{efros2003} \cite{hatun2008} \cite{hu2009} \cite{wong2007} \cite{thurau2008}}
&
\centering{\cite{blank2005} \cite{mokhber2008} \cite{cuntoor2006a} \cite{fathi2008} \cite{ke2005} \cite{ke2007b} \cite{laptev2007a}} & \\
\toprule
\end{tabular}
\end{center}
\label{tab:summary-papers1}
\end{table}

%
\begin{table}[ht]\footnotesize
\caption{Summary of Action Recognition Approaches Based on Global Statistical Representations (table best viewed in color).}
\begin{center}
\begin{tabular}{m{0.15\textwidth} m{0.15\textwidth} m{0.15\textwidth} m{0.15\textwidth} m{0.15\textwidth}}
\toprule
\multicolumn{5}{c}{\textbf{\textcolor{repcolor}{Basic Representation -- Statistics}}}
\\
 \multicolumn{5}{c}{\textit{\textcolor{assumpcolor}{Main Assumption: Global statistics capture relevant information for action recognition.}}}
\\
\multicolumn{5}{c}{[Related Papers by Subcategory (below)]}
\\
\midrule
 \multicolumn{2}{c}{\centering{\textcolor{repcolor}{Low-level Features}}} & \multicolumn{2}{c}{\centering{\textcolor{repcolor}{Local Descriptors}}} & \multicolumn{1}{c}{\centering{\textcolor{repcolor}{Concepts}}}
\\
 \multicolumn{2}{m{0.3\textwidth}}{\centering{\textcolor{assumpcolor}{\textit{Low-level features can indicate actions being performed.}}}}
 & \multicolumn{2}{m{0.3\textwidth}}{\centering{\textcolor{assumpcolor}{\textit{Mid-level local descriptors are better suited to capture relevant information.}}}} &
 \multicolumn{1}{m{0.3\textwidth}}{\centering{\textcolor{assumpcolor}{\textit{Concepts occurrences can indicate the actions being performed.}}}}
 \\
\centering{\textcolor{repcolor}{Domain-Oriented}}
&
\centering{\textcolor{repcolor}{Generic}}
&
\centering{\textcolor{repcolor}{BoVF}}
&
\centering{\textcolor{repcolor}{Other}} &
\\
 \centering{\textcolor{assumpcolor}{\textit{Domain information must guide the choice of relevant low-level features.}}}
 &
 \centering{\textcolor{assumpcolor}{\textit{Generic low-level features are able to capture relevant information.}}}
 &
 \centering{\textcolor{assumpcolor}{\textit{A histogram of quantized local descriptors can be associated with actions.}}}
 &
 \centering{\textcolor{assumpcolor}{\textit{Other ways than BoVF can better capture relevant information from local descriptors.}}} &
 \\
\midrule
\centering{\cite{xie2004} \cite{papadopoulos2008} \cite{tien2008} \cite{snoek2005}}
&
\centering{\cite{lavee2007} \cite{ermis2008} \cite{ma2009} \cite{zelnik-manor2001} \cite{zelnik-manor2006} \cite{shechtman2005} \cite{shechtman2007b} \cite{seo2009} \cite{li2007} \cite{ahammad2007} \cite{wang2007a} \cite{ning2009} \cite{chaudhry2009}}
 &
\centering{\cite{lopes2009b} \cite{schuldt2004} \cite{laptev2007b} \cite{dollar2005} \cite{scovanner2007} \cite{ning2007} \cite{niebles2008} \cite{marszalek2009} \cite{zhao2008} \cite{laptev2008} \cite{liu2008a} \cite{chen2008} \cite{zhou2008}}
&
\centering{\cite{oikonomopoulos2005} \cite{oikonomopoulos2006} \cite{savarese2006} \cite{nowozin2007} \cite{gilbert2008} \cite{ryoo2009} \cite{uemura2008} \cite{sun2009} \cite{messing2009}}
&
\multicolumn{1}{m{0.3\textwidth}}{\centering{\cite{ebadollahi2006} \cite{haubold2007} \cite{xu2008}  \cite{wang2008}}}
 \\
\toprule
\end{tabular}
\end{center}
\label{tab:summary-papers2}
\end{table}

\begin{table}[ht]\footnotesize
\caption{Summary of Hybrid Action Recognition Approaches (table best viewed in color).}
\begin{center}
\begin{tabular}{c}
\toprule
\textbf{\textcolor{repcolor}{Basic Representation -- Hybrid}} \\
\textit{\textcolor{assumpcolor}{Main Assumption: combinations of different approaches produce enhanced recognizers.}} \\
\multicolumn{1}{c}{[Related Papers (below)]} \\
\midrule
\cite{niebles2007} \cite{mikolajczyk2008} \cite{junejo2008} \cite{bregonzio2009} \cite{sun2009b} \cite{oikonomopoulos2009}\\
\toprule
\end{tabular}
\end{center}
\label{tab:summary-papers3}
\end{table}

\section{Approaches Based on Models of the Moving Objects}
\label{sec:object-based}
In this section, approaches which do assume that specific objects are in the scene under constrained conditions are discussed according to the structure proposed in upper part of Figure~\ref{fig:approaches}. This category of approaches relies on models for those objects assumed to be performing the actions, models that can be either explicit (parametric) or implicit.

As indicated in Figure~\ref{fig:approaches}, approaches based on object models are those that appeared first in the literature, out of traditional motion capture research. Some recent and/or classical papers of each branch are cited as needed throughout the following text, but the list presented is not meant to be exhaustive, since those approaches are broadly covered in previous surveys (Section~\ref{sec:previous-surveys}). Therefore, this section is mainly aimed at providing a high-level overview of that category of approaches, detailing it enough to give the reader some perspective on the evolution of the area towards less constrained techniques.

\subsection{Using Parametric Object Models}
The approaches based on parametric object models are the ones more directly related to motion capture techniques. In such approaches, moving objects (e.g., the human body, the hand, cars in a parking area) are assumed to follow a specific model and visual data is matched against that model to infer its parameters. In the action modeling step, different sets of values for the model parameters are then associated with different actions.

Approaches based on parametric models can be split into two major subgroups: those which parameterize the object internal structure, and those which ignore such internal structure and instead, parameterize only objects trajectories.

\subsubsection*{Models of the Objects Internal Structure}
These approaches, which are among the earliest techniques developed for action recognition, are mostly derived from motion capture techniques. Thus, they start by defining a detailed model of the internal structure of a pre-defined moving object and then adjusting the visual data to that model. The most commonly modeled objects are the entire human body as well as body-parts, such as hand models aimed at gesture recognition, for example. A classical example of human-body model (the so-called stick model) can be seen in Figure~\ref{fig:approaches}(a), from \cite{yilmaz2005}.

Strictly speaking, a detailed explicit model would require a great amount of 3D data, like in \cite{yilmaz2005}, thus leading to a high computational complexity. In addition, in most real-world scenarios, 3D data is simply not available. Hence, a number of approaches avoid that requirement by using simplified models. In \cite{peursum2004, peursum2005}, for example, a simplified stick model is obtained from silhouettes. Poses are modeled based on a few points from the body in \cite{jiang2006b}. In \cite{kosta2006} the human body is considered as a cooperative team of agents where each team member is a limb of the body. In \cite{filipovych2008} and \cite{wang2009} constellation models are employed to describe human poses. In \cite{dhillon2009}, body parts are tracked using mixture particle filters and clustering the particles locally.

\subsubsection*{Trajectory Models}
In trajectory-based approaches, the global motion of the objects is considered the only relevant information for action recognition.  In other words, the internal state of the moving objects is ignored and such objects are represented mainly by their position tracked over time. Actions are then modeled by trajectories parameters, which, in turn, can come from a number of different trajectory models. Trajectory-based approaches are very common in surveillance scenarios -- such as the one depicted in Figure~\ref{fig:approaches}(b), from \cite{chen2008}. A number of surveys specifically devoted to them were already discussed in Section~\ref{sec:previous-surveys}.

In \cite{abdelkader2008}, for example, it is argued that activities can be modeled by any representative shape associated with the activity to be modeled, and as an example case, the shape of the trajectories of a set of points associated with the moving object is analyzed. Instead, a large number of proposals analyze the trajectory of a unique point, as in \cite{cuntoor2008a}, which is aimed at identifying common office activities by the analysis of hand trajectories. In \cite{hu2008a} and \cite{hu2008b}, the focus is on detection of abnormal events in crowded scenes. In such scenes, tracking the objects of interest is particularly challenging. To overcome this, global motion fields are analyzed in order to discover \emph{super-tracks}, which are intended to capture predominant motion patterns that are then used to model events.

\subsection{Using Implicit Object Models}
In this class of approaches, the area around the moving object --- like a silhouette or a bounding box, for example --- is detected and submitted to some kind of global description. This line of work assumes that explicit details of the object structures are not necessary for action recognition. Rather, the global features of a \ac{ROI} defined around the object implicitly capture its model, at a lower cost. The lower computational effort offered by this basic idea gave rise to a variety of similar approaches, which can be distinguished between those using 2D templates and those exploiting spatio-temporal 3D templates.

\subsubsection*{Implicit Object Models Based on 2D Templates}
A landmark paper using implicit object modeling as the basic representation is \cite{bobick2001}. In this paper, two different 2D templates computed from extracted silhouettes is proposed: (a)~\ac{MEI}, which is a binary image indicating where the motion occurred during the sequence; (b)~\ac{MHI}, which is a gray level image where brighter pixels indicate the recency of motion (in other words, the brighter the pixel, the more recent the motion occurred there). Both \ac{MEI} and \ac{MHI} images are described by seven Hu moments, which are meant to carry a coarse shape description which is invariant to scale and translation (Figure~\ref{fig:approaches}(c)). The \ac{MEI}/\ac{MHI} representation proposed by \cite{bobick2001} became the basis of a great number of extensions and variations, mostly applied to scenarios with relatively stable backgrounds (like the work of \cite{hu2009} aimed at surveillance applications).

Another classical approach using 2D templates as primary video representations appears in \cite{efros2003}, which addresses the problem of recognizing human actions from medium resolution videos. This approach relies on the detection and stabilization of a bounding box containing the human figure. The description of such boxes is based on the optical flow projected into motion channels, which are blurred with a gaussian filter to reduce the sensitivity to noise which is typical of optical flow estimations. Such motion descriptors are later used by several other authors (see \cite{wang2007a} and \cite{fathi2008}, for instance).

In \cite{wong2007}, the internal part of previously detected motion regions are described by \ac{BoVF}, a statistical representation based on interest points that is further detailed in Section~\ref{sec:bovf-approaches}.

Along with the global spatial descriptors, the information encoded in the sequential nature of video is explicitly taken into account by some authors relying on 2D templates. In \cite{hatun2008}, a bounding box centered at the moving body is described based on radial \ac{HoG}. Such histograms are then clustered to create a codebook of poses, based on which each video is described by two alternative representations, namely: bag-of-poses and sequence-of-poses. A similar approach -- using \ac{NMF} to build a bag-of-poses representation -- is presented in \cite{thurau2008}.

\subsubsection*{Implicit Object Models Based on Spatio-temporal 3D Templates}
In this category, actions are represented as 3D volumes in space-time. Such spatio-temporal volumes are created by aligning and stacking 2D information (e.g., silhouettes, contours, bounding boxes). The exploration of space-time volumes built on silhouettes for action recognition was first proposed in \cite{blank2005}. In their proposal, the properties of the Poisson equation are used to create a representation in which the values reflect the relative position of each internal position in the volume (Figure~\ref{fig:approaches}(d)).

The principle behind such approaches is that spatio-temporal volumes contain both static and dynamic information and are thus  better suitable as representations for action recognition. Similarly to what happened with 2D template-based approaches, the initial idea of \cite{blank2005} is further explored in a number of subsequent papers, which describe the spatio-temporal volume using different techniques. For instance, \cite{mokhber2008}, characterize the spatio-temporal volumes by their 3D geometric moments. In \cite{cuntoor2006a}, it is proposed an algebraic technique for characterizing the topology of those volumes. Finally, the representation used by \cite{fathi2008} can be considered an extension of the work of \cite{efros2003} to a 3D spatio-temporal volume.

In \cite{ke2005} and \cite{ke2007b}, the authors explore space-time volumes at a smaller scale. Videos are over-segmented in space-time, creating micro-volumes, which are described based on optical flow information. They are then compared against manually segmented action templates, by a shape-matching technique adjusted to deal with over-segmentation.

In order to distinguish between drinking and smoking actions, \cite{laptev2007a} use densely sampled \ac{HoG} and \ac{HoF} 3D descriptors computed over manually cropped regions around people's faces, used as input to a cascade of weak classifiers learned by AdaBoost.

\section{ Approaches Based on Global Statistics}
\label{sec:statistical-approaches}
Approaches which rely on the detection of moving objects share the drawback of depending on computer vision tasks -- such as background segmentation and tracking -- which are themselves open research issues. The lack of general solutions to those tasks leads to an excessive number of assumptions about what is in the scene, which ultimately makes such approaches applicable only to very constrained scenarios.

To cope with more realistic and unconstrained settings, different approaches make no assumption on the presence of any specific object in the scene, thus making object detection unnecessary. Instead, those approaches compute global statistics on different data. Statistics on \textit{low-level features} (such as color, texture and optical flow, for example) can be computed either as generic descriptors or guided by specific information about the application domain, as further discussed in Section~\ref{sec:low-level}. Mid-level \textit{Local descriptors} built on low-level data around selected points gave rise to an important branch of approaches based on \acf{BoVF}, which are essentially histograms of quantized local descriptors. BoVF-based approaches are thoroughly discussed in Section~\ref{sec:bovf-approaches}. Although \ac{BoVF}-based approaches dominated the scenario of statistical approaches in recent years, alternative proposals to gather information coming from local descriptors can be found and are discussed in Section~\ref{sec:other-mid-level}. A third research line exploits the probabilities of high-level semantic concepts (e.g., sky, airplane, people) appearing in a video to infer the action taking place in it. These approaches are detailed in Section~\ref{sec:concept-based}

\subsection{Using Statistics of Low-level Features}
\label{sec:low-level}
In this category of action recognition approaches, low-level features of the video are statistically summarized and such summary is used as the video representation. Since the direct usage of low-level features is prone to suffer more intensely the effects of the semantic gap, some authors use previous knowledge about the application domain to guide the choice of features. Generic features without specific links to the application domain have also been exploited, although such approaches tend to be focused on a handful of constrained settings.

\subsubsection{Low-level Statistics Guided by Domain Knowledge}
A combination of low-level features and some previous domain knowledge is common in scenarios where the possible backgrounds are limited in number and have distinct global appearance. In professional sport videos, for instance, camera effects are commonly related to specific events. This fact is used in \cite{xie2004}, in which dominant color ratio and motion intensity are computed to segment soccer videos between play and break intervals.

Another application for approaches based on low-level statistics is explored in \cite{papadopoulos2008}, which use global motion information to identify generic, coarse-grained events in news videos, like \textit{anchor}, \textit{reporting}, \textit{reportage} and \textit{graphics}.

Many approaches based on low-level statistics appear as part of multimodal frameworks (see, e.g., \cite{tien2008}), in which audio-visual features are mixed with high-level information to detect events in tennis games. The full exploration of multimodal frameworks is outside the scope of this paper, although we point the reader to \cite{snoek2005} for a related survey.

\subsubsection{Generic Low-level Statistics}
A variety of approaches for computing generic global statistics based on low-level features, which do not rely on domain information, have been proposed in the recent literature. In most cases, they relate to constrained applications.

In \cite{lavee2007}, a framework aimed at searching for suspicious actions represents candidate video segments by histograms of intensity gradients both in spatial and temporal dimensions, over four different temporal scales.
In \cite{ermis2008}, the typical dynamics of a surveilled environment is captured by statistical analysis of a 2D field containing the maximum activity of pixels.  From that field, a model for normal behavior is produced, allowing comparisons with other videos so as to detect abnormal activities.
Surveillance scenarios are also the focus of \cite{ma2009}, which propose a dynamic texture descriptor based on local binary patterns extracted from the three orthogonal planes formed by the spatial and temporal axes. The videos are represented by sequences of such descriptors computed over subsequent spatio-temporal subvolumes.

In \cite{zelnik-manor2001} and \cite{zelnik-manor2006}, a generic approach for what is called \acf{SFF} is presented. Their approach is based on the absolute values of normalized gradients computed over all space-time points, extracted in a temporal pyramid, to cope with different temporal scales. Points with gradients below a threshold are ignored in order to save time, and the remaining ones are described by the gradient components in $x$, $y$ and $t$ directions, for all temporal scales considered.

Similarly, in \cite{shechtman2005}, underlying motion patterns are applied to identify video segments similar to a query sequence. This is done by computing the correlation of such motion patterns in the query video segment with a larger video sequence.  The peaks in the correlation surface correspond to similar sequences. In this approach, the motion is estimated from the gradients inside small spatio-temporal patches or cuboids, instead of relying on expensive flow computations.

The proposal of \cite{shechtman2007b} computes the similarity between images or videos  by matching local self-similarities. Those are computed at pixel level, taking into account the similarity between a small patch around the considered pixel and a larger region surrounding it.

Also aimed at \ac{SFF} applications, \cite{seo2009} propose to recognize actions from a unique example, by using local regression kernels based on weights computed on the video pixels and their neighbors both in space and time.

In \cite{li2007}, the optical flow computed for the entire video is represented by magnitude and orientation. A histogram is built on the quantized orientation, using only pixels for which the flow magnitude is above a certain threshold. Also, the flow of the considered pixels is weighted by their magnitude. The normalized histograms for the training sequences for each class are submitted to PCA for dimensionality reduction.

In \cite{ahammad2007}, motion vectors from the compressed-domain are used to estimate motion fields, which are then submitted to a hierarchical agglomerative clustering algorithm, in order to create an organizing hierarchy for videos which are presumed to be based on actions.

In contrast to the \ac{BoVF}-based approaches (see Section~\ref{sec:bovf-approaches} below), \cite{wang2007a} argue that human actions should be characterized by large-scale features instead of local patches. Therefore, the authors consider the frame as the basic unit for initial description, which is made in terms of the motion descriptors proposed by Efros in \cite{efros2003}. Their ``visual vocabulary'' is then built on those global frame features, whose space is quantized by the k-medoid clustering algorithm. Finally, each video sequence is represented in terms of the frequency of such ``frame-words".

In \cite{ning2009}, a hierarchical space-time model is implemented in two layers: the bottom layer of features composed of a bank of 3D Gabor filters; the second layer in the proposed hierarchy are histograms of Gabor orientations. This proposal is based on that of \cite{serre2007} for object recognition, which tries to mimic organic visual systems, which are seen as being composed of two kinds of brain cells with different roles in the recognition process.

Finally, in \cite{chaudhry2009}, the temporal evolution of \acf{HoF} features gathered from each frame are modeled by a non-linear dynamical system.

\subsection{Approaches Based on Statistics of Local Descriptors}
Last section discussed the first attempts at avoiding the constraints imposed by object models for action recognition. However, most of those approaches either rely on domain knowledge or are focused on constrained settings or databases, like surveillance or \acf{SFF} applications. Another drawback of those approaches is that, being based on dense low-level features, they demand great computational effort.

To mitigate those drawbacks, approaches based on mid-level local descriptors, mostly computed on a (potentially small) number of interest points emerged as a promising trend for action recognition. More specifically, approaches based on histograms of quantized local descriptors -- known as \acf{BoVF}\footnote{Due to the lack of standard terminology, those approaches have also been denominated bag of visual words, bag of keypoints, bag of features or bag of words in the literature.} -- have shown to provide consistently good results reported by a number of independent authors in a variety of scenarios, including datasets composed of professional and amateur realistic videos.

Despite the success of \ac{BoVF}-based approaches, there are a few other strategies to gather information from local descriptors. These alternative strategies are gathered in a separate category in the proposed framework.

\subsubsection{Using \acf{BoVF}}
\label{sec:bovf-approaches}
\ac{BoVF} representations are inspired by traditional textual \ac{IR} techniques, in which the feature vectors that represent each text document in a collection are histograms of word occurrences \cite{baeza-yates1999}\footnote{In fact, each histogram bin reflects not a single word, but a family of words represented by their roots.}. Such representation is referred to as \acf{BoW}, in order to emphasize that it is comprised of orderless features.

A remarkable difference in the analogy between \ac{BoVF}s and \ac{BoW}s is the need to define what constitutes a \textit{visual word}. Such ``definition'' is achieved in practice by a process called \textit{vocabulary (}or \textit{codebook) learning}, consisting of the quantization of the descriptors' feature space, typically computed by clustering. A detailed introduction on how BoVF representations are build both for images and videos can be found in \cite{lopes2009b}.

BoVF-based approaches have been first applied to object classification and have proved very robust to background clutter, occlusion and scale changes, indicating their potential for challenging object recognition settings (\cite{agarwal2004}, \cite{lazebnik2006}, \cite{zhang2007}, \cite{jiang2007}, \cite{wong2007b}, \cite{sun2009}). \ac{BoVF} and its variations have demonstrated similar strengths when applied to action recognition, thus becoming, by far, the most common base representation found on recent proposals.

The relevance of \ac{BoVF}-based action recognition is reinforced by the fact that those schemes became a common testbed for several spatio-temporal points detection and description algorithms. The work of \cite{schindlerGrant2008}, for example, compares different alternatives for interest point detectors/descriptors applied in a classic \ac{BoVF} representation for Internet videos. Similar comparisons can also be found in \cite{wong2007b}, \cite{klaser2008}, \cite{willems2008}, \cite{kaiser2008} and \cite{rapantzikos2009}.

To the best of our knowledge, the approach proposed by \cite{schuldt2004} is the seminal work on \ac{BoVF} techniques applied to action recognition. For the low-level features, the spatio-temporal interest points proposed in \cite{laptev2003} are described by spatio-temporal jets. K-means clustering is applied to create the quantized vocabulary, based on which the histogram of local features is computed. This is also the work which introduced the \ac{KTH} action database, which later became a \emph{de facto} standard for action recognition algorithms. The work described in \cite{schuldt2004} is extended in \cite{laptev2007b}, which proposes a mechanism for local velocity adaptation aimed at compensation of camera motion that could affect local measurements.

The work of \cite{dollar2005} extends previous work on object recognition based on sparse sets of feature points. The interest points selection method applied in this work is based on separable linear filters.  Three descriptors for the cuboids delimited around the selected points are tested: \emph{normalized pixel values}, \emph{brightness gradients} and a \emph{windowed optical flow}. \ac{PCA} is used for dimensionality reduction of the point descriptors and a typical \ac{BoVF} signature is then built on them. The k-means clustering algorithm is used for defining the dictionary.

Another BoVF approach is proposed by \cite{scovanner2007}, this time based on an extension of the \ac{SIFT} descriptor \cite{lowe1999}. The new descriptor adds temporal information,  extending the original \ac{SIFT} descriptor to a 3D space. Instead of using the \ac{SIFT} detector, points are selected at random. Histograms built on a codebook created with k-means are the initial signatures. Then, to create an enhanced representation, a criteria based on the co-occurrence of visual words is applied to reduce the vector dimension. In other words, those visual words which co-occur above a certain threshold are joined.

In \cite{ning2007} the local descriptors are based on the responses to a bank of 3D Gabor Filters, followed by a MAX-like operation. Such features are computed on patches delimited by a sliding window and described by histograms generated by the quantization of the orientations in nine directions. The quantization of those histograms into a codebook is learned from a gaussian mixture model.

In \cite{niebles2008}, a \ac{BoVF} representation based on the features proposed by Dollar \cite{dollar2005} is used together with generative models -- unlike previously discussed methods which are based on discriminative ones -- for action recognition. Two methods borrowed from traditional textual Information Retrieval research are examined: \ac{pLSA} and \ac{LDA}.

In \cite{marszalek2009},  a joint framework using \ac{BoVF}s both for scene and actions recognition is proposed. The underlying assumption is that scenes can provide contextual information to improve the recognition of some actions classes. Initially, movie scripts provide for automatic annotation of scenes and actions. Text mining is then applied to discover co-occurrences between scenes and actions. Finally, separate \ac{SVM} models based on \ac{BoVF}s are learned using the same approach described in \cite{laptev2008}.
%
\subsubsection*{\ac{BoVF} Variations}
A number of variations over the typical BoVF scheme have been proposed, mostly aimed at dealing with specific recognized limitations of pure \ac{BoVF}-based approaches, the main ones being the lack of structural information and the poor quality of the visual vocabulary.

In \cite{zhao2008}, the lack of structural information of \ac{BoVF} representations is addressed in a rather direct manner: each frame is subdivided into cells, over which \ac{BoVF} based on Dollar's features \cite{dollar2005} are computed. Additionally, motion features from neighbor frames are used in a weighted scheme which takes into account the distance of the neighbor from the actual frame. A spatial-pyramid matching, similar to the one used in \cite{lazebnik2006}, is then applied to compute the similarity between frames. Finally, frames are classified individually, and a voting scheme decides the final video classification.

In \cite{laptev2008}, it is proposed a \ac{BoVF} representation built from the \ac{STIP} presented in \cite{laptev2003}, but without scale selection. The points are described by \acf{HoG} and \acf{HoF}, computed over the spatio-temporal volumes positioned around them. To add some structural information, each video volume is subdivided by a grid, so that at recognition time different configurations for the grids are considered, using a multi-channel \ac{SVM} classifier.

Most authors working on BoVF-based approaches for actions recognition learn the vocabulary by using the k-means clustering preceded by a \ac{PCA}-based dimensionality reduction. Nevertheless, some alternatives to vocabulary learning have been proposed, either by enhancing the vocabulary delivered by k-means or by applying alternative clustering techniques.

The work of \cite{liu2008a} merges k-means results to produce an enhanced vocabulary, indicating that better vocabularies can have a significant impact on recognition. Using features similar to those of Dollar \cite{dollar2005}, they propose a \ac{MMI} criteria to merge the cuboid clusters output by k-means. Those new clusters are called \ac{VWC}. Additionally, two approaches to add structural information are explored: spatial correlogram, with the distances quantized in a few levels and the \ac{STPM} of \cite{lazebnik2006}.

Another example of the impact of  a better vocabulary is found in \cite{jiang2009}, where the authors proposed an enhanced BoVF in which the relationships among visual words are explored. This is pursued with the aid of a visual ontology inspired on WordNet \cite{wordnet}, a textual ontology extensively applied for text retrieval. Their visual ontology is built by applying agglomerative clustering to the visual words previously discovered by k-means. From that, it is possible to compute the specificity (i.e., the depth in the ontology tree), path length (i.e., the number of links in the path between two words) and information content (relative to the probability of word occurrences). Those precomputed values are used in a soft-weighting scheme aiming at better evaluating the significance of each word.

In \cite{liu2009}, a BoVF combining dynamic and static local features is proposed to address action recognition in YouTube videos. Static information captured by three different interest point detectors are described by SIFT descriptors. Motion is collected by Dollar's interest points \cite{dollar2005}, described by gradient vectors. The spatio-temporal distribution of motion features is used to localize coarse \acp{ROI}, which are used together with the PageRank algorithm, for prunning spurious features. In addition to such motion-guided feature selection, the authors propose a procedure to create a semantic visual vocabulary, which involves enhancing the result of k-means by using a technique based on the KL-divergence algorithm.

In \cite{chen2008}, a modified version of k-means which takes into account the spatial localization of the interest points is applied to form the codebook. Additionally, a 3D extension of the Harris detector alternative to that of Laptev \cite{laptev2003} is proposed, aimed at selecting a denser sampling. Cuboids defined around the detected points are described in terms of shape and motion.

Finally, in \cite{zhou2008}, the visual word distribution is described by \ac{GMM} of SIFT descriptors. These \ac{GMM}s are specialized for each video clip, and a kernel for video comparison is built on the Kullback-Leibler divergence (\cite{moreno2003} \textit{apud} \cite{zhou2008}).

\subsubsection{Alternative Strategies to Capture Information from Local Descriptors}
\label{sec:other-mid-level}
Despite the great success of BoVF-based approaches and their variations, achieving high recognition rates in truly realistic databases remains an open challenge. A number of authors have been trying alternative ways to collect relevant information out of local descriptors.

In \cite{oikonomopoulos2005}, the salient-points detected at peaks of activity are clustered into salient regions, whose scales are correlated to the motion magnitude. Noisy interest points are discarded, so the videos are described by remaining points inside detected salient regions. Since such representations do not have the same dimension for all video sequences, the Chamfer distance is used as the kernel for a \ac{RVM} classification algorithm. A space-time warping adjustment scheme is applied to deal with varied execution speeds. Some variations to this overall scheme are presented in \cite{oikonomopoulos2006}.

In \cite{savarese2006}, it is argued that correlograms can capture the spatial arrangement of codewords in the case of object classification. This is applied in \cite{savarese2008}, in which an extension of the spatial correlatons (quantized in \emph{correlograms}) is proposed for action recognition. Rather than building a histogram of interest-points, as in typical BoVF approaches, action is modeled as a collection of space-time interest points where each interest point has a label from the vocabulary of video-words. So, the video sequences are composed of sets of video words and their spatio-temporal relations are described in form of spatio-temporal correlatons. Actions are modeled by estimating -- with \ac{pLSA} -- the codewords distribution for each particular class.

In \cite{nowozin2007}, it is observed that the lack of structural information also means the absence of temporal sequencing. The representation proposed uses Dollar's features \cite{dollar2005}, but the histogram built on them is made up of temporal bins. PrefixSpan algorithm is used for mining frequent sequences and the LPBoost algorithm is used to identify the most discriminative ones.

In \cite{gilbert2008}, dense corner features are hierarchically grouped in space and time to produce a compound feature set. Data mining is applied to group features in multiple stages, from the initial low-level features until a higher level in which the relative positions of groupings are used. As in \cite{lopes2009b}, 2D features are collected from the  planes defined by the coordinates $(x,y)$ -- the frames -- $(x,t)$ and $(x,t)$, which are, in the former case, considered as distinct channels. Motion is captured by dominant orientation and points are described only by their scale, corresponding channel and orientation, instead of more complex descriptors like \ac{SIFT} or \ac{STIP}. Transaction vectors are built based on neighbor interest points and the Apriori algorithm is applied to the transactions in order to find association rules between transactions vectors and actions.

In \cite{ryoo2009}, a matching kernel function for comparing spatio-temporal relationships among interest points is proposed for detection and localization of multiple actions and interactions in unsegmented videos. First, interest points are detected and described as usual. Afterwards, pairwise relationship predicates are used to describe the structural relations. Temporal relations are described by Allen's taxonomy (\textit{equals}, \textit{before}, \textit{meets}, \textit{overlaps}, \textit{during}, \textit{starts} and \textit{finishes}) \cite{allen1994} \textit{apud} \cite{ryoo2009}, with respect to the interval limits given by the volume patch dimensions projected onto the temporal axis. Similar spatial predicates are created, so that temporal and spatial 3D relationship histograms aimed at capturing both appearance and point relationships in the video can be computed. Finally, the proposed matching kernel captures the similarity between two histograms by their intersection.

The work presented in \cite{uemura2008} proposes a video representation in the form of a vocabulary-tree, based on the outputs of several interest point detectors plus a dense sampling for action recognition and localization. Motion compensation is achieved by using previously tracked features to perform the segmentation of the image into motion planes. Such a segmentation is performed by an initial color-based segmentation followed by homography computation using RANSAC. The homography is then used to correct the motion of the features inside each dominant plane.

The fact that humans can recognize actions just by observing some tracked points has been explored in several approaches relying on trajectory models of points placed at specific body parts. The work of \cite{sun2009} extends this notion into a more generalizable approach, in which the authors propose to gather information about the spatio-temporal context of tracked SIFT points in a hierarchical three-level scheme. In the first level -- the point-level context -- local statistics of gradients along point trajectories are computed. In the second level -- the intra-trajectory context -- the dynamic aspects of those trajectories in the spatio-temporal space are considered. Finally, in the coarser level -- the inter-trajectory context -- the information about the spatio-temporal co-ocurrences of trajectories distributions is collected.

In a similar vein, the tracks of a set of features -- detected and tracked by the algorithm proposed in \cite{lucas1981}, but with weaker constraints -- is employed by \cite{messing2009}. Such trajectories are described by the history of their quantized velocities.

\subsection{Using Concept Probabilities}
\label{sec:concept-based}
Using similar ideas from \ac{CBVR} systems \cite{snoek2008}, some authors have proposed to use higher level concepts as the building blocks of video representations aimed at action recognition.

In \cite{ebadollahi2006}, 39 semantic concept detectors from \ac{LSCOM}-lite -- \ac{SVM} classifiers based on raw color and texture features \cite{naphade2005} -- are applied to video I-frames. Then, the trajectories of those concepts in the concept space are analyzed by \ac{HMM}, one for each concept axis. Their work reinforces the results of \cite{kennedy2006}, providing additional evidence that, for some concepts, the dynamic information is essential. The authors found that dynamic information enhanced recognition results of the following concepts: \emph{riot}, \emph{exiting car} and \emph{helicopter hovering}.

In \cite{haubold2007}, \ac{MPEG} motion vectors are summarized in a motion image which describes the global motion pattern of a video shot. Motion images are combined with color and texture features and used as input for several weak \ac{SVM} classifiers. The output of such classifiers are fused together to compose the video feature vector. The approach is tested on the \ac{TRECVID}-2005 dataset, for selected dynamic concepts only, comparing favorably with results based on motion direction histograms and motion magnitude histograms.

In \cite{xu2008}, 374 concepts are selected from the \ac{LSCOM} ontology to be detected by three different \ac{SVM} detectors based on histograms of low-level features (grid color moments, Gabor textures and edge direction histograms). The results of those classifiers are fused together in order to produce scores for each concept. Variations in the duration of action clips are dealt with by applying the \ac{EMD} in multiple temporal scales.

A framework for event detection presented in \cite{wang2008} starts by the application of \ac{BoVF}-based approach to detect a number of concepts. Relative motion of keypoints between successive frames is used to aid the spatial clustering of visual words. A visual word ontology is then built based on the output of the spatial clustering, in order to take into account the correlation of visual words potentially related to the same object or object parts. The final representation is a collection of \ac{BoVF}s built on those roughly segmented regions.

\section{Hybrid Approaches}
This section is devoted to action recognition approaches that fuse information coming from both object models and global statistical representations. It is worth noticing that, from the point of view of generalization ability, such mixed approaches are limited by the representation whose computation imposes stronger constraints.

In \cite{niebles2007}, it is proposed a hybrid approach where constellation models are used to add geometric information to the classical BoVF representation. This is done by modeling actions within a two-layered hierarchical model. The higher layer is comprised of selected body parts, which are then described as \ac{BoVF}s. The \ac{BoVF}-based system proposed is based on Dollar's features \cite{dollar2005}, together with sampled edge points described by shape context.

The work presented in \cite{mikolajczyk2008} mixes low-level, local descriptors and shape-based representations. They build several vocabulary trees on points selected by five different 2D point selectors. To include dynamic information, motion maps are obtained from a Lucas-Kanade optical flow computation, and motion is represented by velocity maps between pairs of frames. A technique for compensation of camera motion based on a global similarity transformation is also presented and applied. A star-shape model -- aimed at coarsely capturing some structural information of the moving object -- is used to guide the process.

In \cite{junejo2008}, the concept of \ac{SSM} is introduced to build video representations aimed at action recognition. An \ac{SSM} is a table of distances between all video frames. Although this definition can be applied for any feature type, in \cite{junejo2008}, they are computed over trajectories of human joints, which are fused together with those computed from \ac{HoG} and \ac{HoF} features. The final descriptor is obtained by considering the \ac{SSM} sequences as images and splitting them into patches, which are described by histograms of gradient directions.

The proposal of \cite{bregonzio2009} works on clouds of interest points collected over different time scales. The distribution of such clouds in both space and time is described by global features. To compose the clouds, they propose a spatio-temporal interest point detector that collects dense samplings of interest points. In order to avoid too many spurious point detections, the moving object is coarsely separated from the background.

In \cite{sun2009b}, \ac{BoVF}s of 2D and 3D SIFT feature descriptors, extracted on 2D SIFT interest points, as well as Zernike moments, are applied to both frames and \ac{MEI} images. The extraction of those features are guided by frame subtraction, which provides a coarse motion-based segmentation. Feature fusion is achieved by simple concatenation of the resulting four descriptors.

The work of \cite{oikonomopoulos2009} proposes another hybrid approach for both recognition and detection of actions in unsegmented videos. Visual codebooks are class-specific and take co-occurrences of visual words pairs into account. The positions of these visual words in relation to the object center are used to model the actions, which therefore implies the need for object segmentation. In addition, spatio-temporal scale adjustments are done manually for training. Finally, a framework for voting over time, which is based on optical flow and appearance descriptors, is proposed for action segmentation.

\section{Summary and Discussion}
\label{sec:discussion}
Figure~\ref{fig:approaches} together with Tables~\ref{tab:summary-papers1},~\ref{tab:summary-papers2}~and~\ref{tab:summary-papers3} summarize our survey on previous and state-of-the-art action recognition approaches. As already discussed throughout the text, approaches relying on object models to describe video content have the drawback of imposing a number of constraints on the action scenario. Such constraints are rarely met in feature movies or user-generated videos found in video sharing systems (e.g., YouTube), thus pushing the research in action recognition towards more general approaches. It is important to notice, though, that some approaches based on object models have proved successful in realistic but restrict application domains, like surveillance and \ac{HCI}, for example. In fact, provided that their constraints can be guaranteed, those approaches should be considered as potential choices in those cases where real-time processing is a requirement. In particular, approaches based on implicit models built on 2D descriptors -- for instance, like those based on \ac{MEI}s and \ac{MHI}s \cite{bobick2001} -- tend to be quite efficient. In addition, the advances in the state-of-the-art of pre-processing techniques like segmentation and tracking can turn some approaches based on object models better suited for realistic environments, possibly giving rise to new hybrid approaches.

Regarding approaches based on global statistics, a clear tendency towards the usage of BoVF representations emerges, specially in those attempts to deal with unconstrained video databases.

Nevertheless, in spite of their success, a number of limitations are yet to be overcome by BoVF-based systems in order to achieve solutions that are mature enough to be incorporated in real-world tools.
One of those issues is the \textit{ad-hoc} nature of the visual vocabulary learning process. Although some papers discussed (Section~\ref{sec:bovf-approaches}) indicate better alternatives to pure k-means, there is no principled methodology neither to build an optimal vocabulary, nor even to preemptively assess the vocabulary quality, given a specific database.

The relatively small number of proposals dealing with local descriptors in alternative ways (Section~\ref{sec:other-mid-level}) prevents the anticipation of specific trends in this direction, but it is worth noticing that both BoVF and non-BoVF-based approaches have been moved from an initial preference for sparse set of interest points to a current trend towards denser sets, on the assumption that they are more appropriate for realistic scenarios. This premise is reinforced, for example, by the comparison among several 3D point detectors and descriptors performed by \cite{wang2009}, which concluded that, except for the \ac{KTH} database (which has very little contextual information, given its neutral background), regularly spaced dense samplings of points perform better at action recognition than interest points.

Approaches based on concept probabilities are somewhat underexplored for action recognition, if one considers their widespread usage in \ac{CBVR} systems. Reasons for this apparent lack of interest might be the scarcity of annotated training samples for each concept classifier, as well as the issues raised by the usage of meta-classification and classification fusion.

The scarcity of labeled data for action recognition itself is an important issue which needs to be tackled in order to allow more significant advances in the area. Some initiatives have generated \textit{de facto} standard databases, like Weizmann \cite{blank2005} and KTH \cite{schuldt2004} controlled databases, and, recently, more  realistic databases like the Hollywood Movies Dataset (\cite{laptev2008}, \cite{marszalek2009}) and the Action Dataset \cite{liu2009}. Such annotation efforts have been fundamental to the research advances in the last few years, but they suffer the limitation of being somewhat isolated efforts and therefore, necessarily limited in size and scope. The \ac{TRECVID} benchmark, which is well-known for its collective annotation efforts at larger scales, has had an event detection/recognition track for a few years, but videos and ground-truth data are available for participants only.

Besides publishing their own databases, some authors propose alternative ways for dealing with the issue of lack of annotated data. In \cite{duchenne2009}, it is proposed a semi-automatic annotation technique based on movies scripts and subtitles. In \cite{ulges2009}, the issue of collecting a large-enough amount of training data for high-level video retrieval is addressed by the usage of videos collected from YouTube filtered by pre-defined categories and tags.

Semi-supervised methods can also be used to produce larger amounts of annotated data from a small number of samples. In \cite{yao2009}, a template is manually generated by cropping bounding boxes from one action example in a clean, controlled database and then applied to detect other instances of the same action in cluttered videos, using \ac{DTW} to deal with variations in the duration of the action. Similarly, in \cite{lin2009}, action prototypes are learned on a laboratory dataset and then applied to action classification in videos with dynamic backgrounds.

Although most current approaches for action recognition focus on enhancing recognition capability, it is worth mentioning that in order to scale up to the dimensions of the web or of any large database, efficiency needs to be taken into account. Some efforts in this direction are available in the literature. In the work of \cite{lin2009}, for example, lookup tables of action prototypes are used to speed-up action classification. In \cite{uemura2008}, \cite{ji2008} and \cite{mikolajczyk2008}, the features are quantized using vocabulary trees, which lead to more efficient matching when compared with traditional codebooks. In the realm of more typical BoVF schemes, compact vocabularies -- like those in \cite{liu2008a}, \cite{jiang2009} and \cite{liu2009} -- can help to reduce the overall computational effort. In \cite{ke2005}, \cite{ke2007} and \cite{laptev2007a}, efficient boosting algorithms are applied for volumetric matching.  In \cite{mikolajczyk2008}, the high computational complexity of the proposed approach -- based on several dense interest point detectors -- is explicitly pointed out, and leaded the authors to implement their recognition framework in a parallel architecture.

\section{Concluding Remarks}
\label{sec:conclu-approaches}
This survey attempts to summarize the efforts of the academic community at the task of recognizing human actions from videos, with emphasis on recent approaches. It proposes a new organizing framework, based on the representations chosen, and therefore, on their underlying assumptions. This organization allows to categorize the newest approaches smoothly alongside the more traditional ones. It also allows to compare and contrast different methods based on their constraints, which, we hope, enables a principled selection of a method, given the application domain. We observe that there is a correlation between our classification criteria and the chronology of methods, indicating a trend toward progressively weakening the constraints imposed on video content.


The greater focus on \ac{BoVF}-based approaches emerges naturally from their potential on the field, making it a promising direction to pursue in the search for effective solutions for recognizing human actions in scenarios of realistic videos. Many questions, however, remain open, and a better assessment of the capabilities of current methods in very challenging scenarios will depend on a collective effort of generating annotated data.


\section*{Acknowledgements}
The authors thank Dr. Ivan Laptev for his comments on earlier versions of this manuscript, as well as the Brazilian funding agencies CAPES, CNPq and FAPEMIG.    
\bibliographystyle{elsarticle-num}
\bibliography{humanActions}







\end{document}